# Data extraction and processing methods to aid the study of driving behaviors at intersections in naturalistic driving

*Shrinivas Pundlik[1,2], Seonggyu Choe[1,2], Patrick Baker[1,2], Chen-Yuan Lee[1,2], Naser Al-Madi[3], Alex R. Bowers[1,2], Gang Luo[1,2]*
[1]Schepens Eye Research Institute of Mass Eye & Ear, Boston MA
[2]Harvard Medical School Department of Ophthalmology, Boston MA
[3]Colby College, Waterville, ME

***Summary:*** Naturalistic driving studies use devices in participants' own vehicles to record daily driving over many months. They offer key insights about real-world driver behavior compared to simulated driving studies; however, due to diverse and extensive amounts of data recorded, automated processing is necessary. This report describes methods to extract and characterize driver head scans at intersections from data collected from an in-car recording system that logged vehicle speed, GPS location, scene videos, and cabin videos. First, from the drive log, intersection locations were marked and the GPS locations and video timestamps were assigned to the selected intersections. Then, cabin and scene videos of the intersections segments were clipped from the recorded data. Custom tools were developed to mark the intersections, synchronize location and video data, and clip the cabin and scene videos for ±100 meters from the intersection location. A custom-developed head pose detection AI model for wide angle head turns was run on the cabin videos to estimate the driver head pose, from which head scans >20° were computed in the horizontal direction. The scene videos were processed using a YOLO object detection model to detect traffic lights, stop signs, pedestrians, and other vehicles on the road. Turning maneuvers were independently detected using vehicle self-motion patterns. Stop lines on the road surface were detected using changing intensity patterns over time as the vehicle moved. The information obtained from processing the scene videos, along with the speed data was used in a rule-based algorithm to infer the intersection type, maneuver, and bounds (entry and exit points of the vehicle for any given intersection). Head scans just before entering and within the intersection were identified. Wide head pose accuracy was evaluated in a controlled experiment for head rotations over the range of ±135°, with a mean absolute error of 6.75±5.0° and an RMSE of 8.4°. We processed 190 intersections from 3 vehicles driven in cities and suburban areas from Massachusetts and California. Ground truth points of vehicle entry and exit into the intersection, as well as the type of intersection traffic control device (traffic light, stop sign, none) and turn maneuver were marked manually for comparison. The automated video processing algorithm correctly detected intersection signage and maneuvers in 100% and 94% of instances, respectively. The median[IQR] error in detecting vehicle entry into the intersection was 1.1[0.4-4.9]meters and 0.2[0.1-0.54] seconds. The median overlap between ground truth and estimated intersection bounds was 0.88[0.82-0.93]. Automating the detection of precise intersection bounds and wide head scan detection makes it feasible to efficiently process large amounts of diverse data and improve reliability of the measurement of behavioral outcomes.

## 1. Introduction

Navigating intersections is a key part of everyday driving; however, about 25% of fatal traffic crashes occur at intersections.[1] It is, therefore, important to study driving behaviors before and within intersections to understand more about factors that may contribute to crash risk. Fundamental questions include whether the driver scanned (looked) to check for hazards before entering the intersection and whether they scanned again after entering the intersection to check for hazards that might suddenly materialize during the intersection maneuver. Scanning is especially important in the context of drivers with visual field deficits who need to scan to see areas of the intersection which would otherwise be obscured by their field loss. For example, at a T-intersection, a driver with hemianopia (the loss of either the left or the right side of the field in both eyes) would have to scan laterally almost 90° to see all of the intersection on the side of the field loss, requiring substantial head rotation.[2] To date, most behavioral data regarding scanning at intersections by people with visual field deficits come from driving simulator studies, where driving can be studied in controlled scenarios.[3-5] However, driving simulators do not replicate all the complexities of the traffic situations and environmental conditions encountered in real-world driving. Therefore, studying real-world driving behavior could provide critical insight into a variety of behavioral, traffic safety and vision science questions related to driving with visual field loss.

Naturalistic driving studies involve analysis of driving (driver) behavior in real-world, everyday driving recorded with an in-vehicle device over many months. Compared to other methods of understanding driving behavior, findings of a well-designed and executed naturalistic study are more likely to be generalizable and closer to the reality.[6] The improved generalizability of naturalistic studies comes with the tradeoff of increased complexity of data acquisition, processing, and analysis. In order to arrive at meaningful conclusions, more data may need to be observed in naturalistic settings compared to simulator studies. Consequently, naturalistic driving studies also tend to be extensive in scope regarding the amount and diversity of the data involved. Therefore, custom methods and tools need to be developed to acquire, process, reduce, and analyze the data for a meaningful understanding of driver behavior.[7] In this document, we describe the technical details about the data acquisition and processing steps we developed to aid in analysis of driver behavior at intersections in naturalistic driving. We evaluate the effectiveness of the data processing methods from real-world recorded data and provide quantitative measures wherever applicable to demonstrate the effectiveness of our methods.

## 2. Naturalistic Driving Data

The data used in this project were collected as a part of a multisite naturalistic driving project evaluating head scanning at intersections of drivers with visual field loss compared to drivers without field loss. The research adhered to the principles of the Declaration of Helsinki, and all participants provided written informed consent before participation. The study protocol was approved by the Institutional Review Board at Mass General Brigham.

Driving data were recorded using a commercial fleet-management telematics solution by Zubie (Bloomington MN, USA), consisting of a Raven+ dashcam recording device (Figure 1; Ottawa ON, Canada) with a road-facing wide dynamic range scene camera (1920×1080, 30 Hz, 140° field of view) to capture the scene through the windshield and a high dynamic range driver-facing cabin camera (1280×720, 30Hz, 130° field of view) to record the driver, among other on-board hardware. The cabin camera had an IR switcher and LED to facilitate nighttime recording of driver activity. We added a physical shutter for the cabin camera which could block the camera from recording inside the cabin (e.g., if the driver did not want video to be recorded or there was a child under 18 years of age in the car). The dashcam was securely placed on the vehicle dashboard or windshield by a member of the research team in a position which did not obstruct the driver's view through the windshield (Figure 1).

The dashcam was connected to the vehicle OBD2 port, from which the vehicle speed was obtained and embedded in the cabin and scene camera video frames. The device recorded the GPS location at a frequency of about 0.5 Hz (once every 2 seconds). Recording started as soon as the ignition was turned on and stopped when the ignition was turned off. All data and videos from both cameras were recorded in an on-board storage card. Data were monitored remotely on a weekly basis to check the dashcam was functioning properly, confirm that video was being captured, and verify that the recorded data had not exceeded the capacity of the on-board storage card. When the storage card was close to being full the participant returned to the study site for the dashcam to be removed or for a new storage card to be installed. Data were transferred from the storage cards to secure network servers for further processing and analysis.

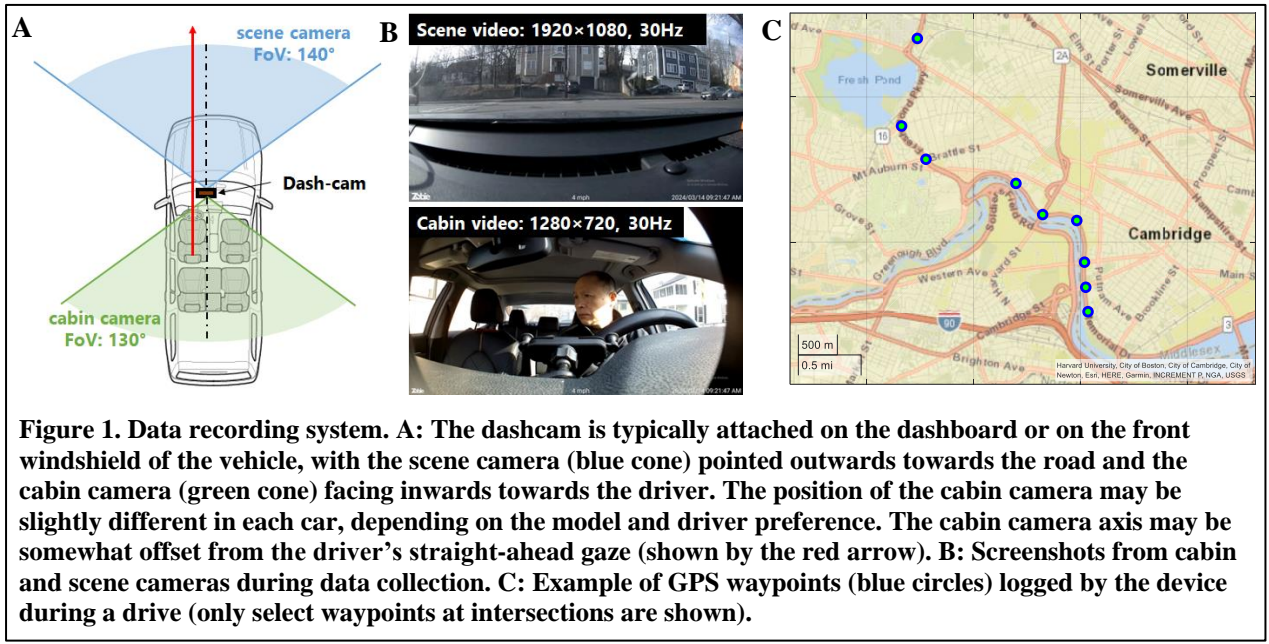

**Figure 1. Data recording system. A:** The dashcam is typically attached on the dashboard or on the front windshield of the vehicle, with the scene camera (blue cone) pointed outwards towards the road and the cabin camera (green cone) facing inwards towards the driver. The position of the cabin camera may be slightly different in each car, depending on the model and driver preference. The cabin camera axis may be somewhat offset from the driver's straight-ahead gaze (shown by the red arrow). **B:** Screenshots from cabin and scene cameras during data collection. **C:** Example of GPS waypoints (blue circles) logged by the device during a drive (only select waypoints at intersections are shown).

## 3. Problem Statement & Solution Framework

Simply stated, our end goal is to analyze real-world driving behavior, especially head scanning at intersections. However, given the large amount of data, anticipated to be typically of the order of tens of hours over hundreds of miles per participant, there is a great need for data reduction. The recording system logs a large amount of data when the vehicle is not at an intersection (for example, highway driving), which while interesting in the long-term, is not relevant to the immediate study goals. Thus data need to be filtered to only retain intersections. Even then, it is not possible to visually inspect, observe, and annotate the various details of navigating an intersection because a driver can easily visit thousands of intersections during the course of their study participation.

In uncontrolled naturalistic settings, this means there are multiple aspects to the problem: i) precisely determining the intersection location in terms of time and distance from a large amount of on-road driving data, ii) detecting head scans and selecting only those which occur within specific intersection regions (e.g., the 20 meters prior to the vehicle entry point), iii) characterizing the intersection – such as the type of traffic control device ("signage"), geometry, maneuver etc., and iv) characterizing the scene such as the traffic density, time of the day, weather etc. Thus, the desired outputs that we require in this application are driver behavior within the cabin (head scanning), behavior manifested in driving (speed, acceleration/deceleration, stops, distance travelled etc.), nature of the intersections navigated (type, geometry, maneuver etc.), and other characteristics of the intersection scenarios (traffic density, visibility, adverse weather, etc.).

Head orientation and hence head scanning could be obtained by processing the cabin videos (section 4.6). The next big question was what information could be used to detect intersection segments from the stored driving data. Specifically, we wanted to detect when the vehicle entered and exited an intersection in addition to the vehicle maneuver, intersection type, and geometry (Figure 2A). One possible option was to use vehicle geolocation data (logged as waypoints every 2 seconds) along with vehicle speed (from OBD2 port), with some help from mapping services (for example, Google Maps API or OpenMaps), to obtain intersection points on the map. Then a zone of fixed radius could be defined around the detected intersection locations to focus on for further analysis. However, there were some major limitations of a geolocation-only strategy in our application: i) GPS location was not accurate – the error could be over many meters, ii) there was some variation in the frequency of waypoint logging (so it was not exactly 2s), iii) there were delays in vehicle speed update affecting computation of distance travelled, and iv) intersection locations and their characteristics (such as signage type) were not reliably and clearly marked in mapping services. These issues led to large errors in estimating the intersection location (Figure 2B&C), and therefore entry and exit points could not be determined precisely.

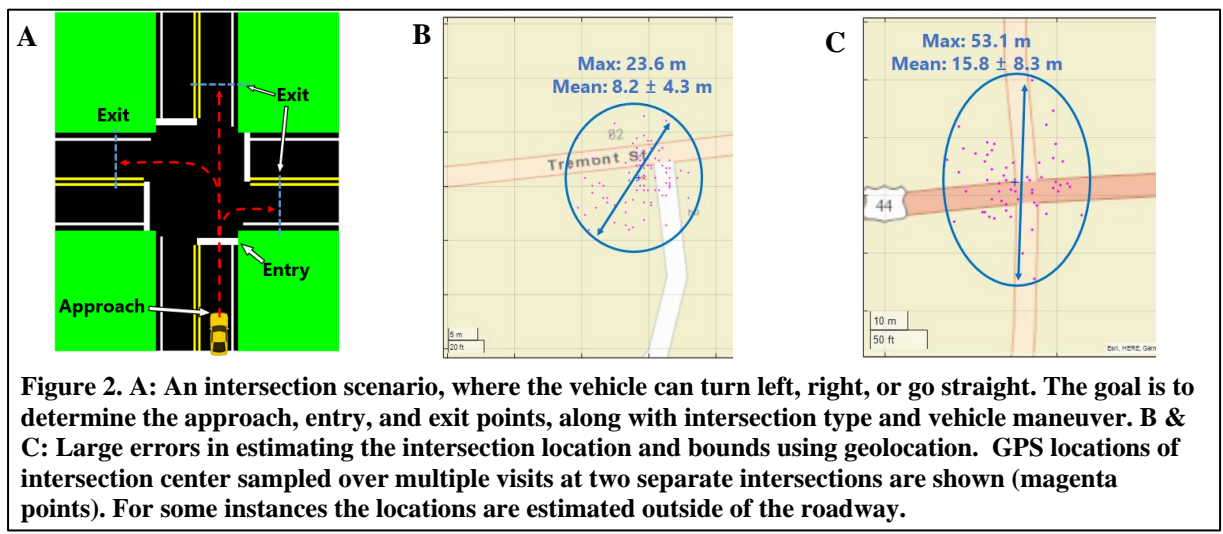

**Figure 2. A: An intersection scenario, where the vehicle can turn left, right, or go straight. The goal is to determine the approach, entry, and exit points, along with intersection type and vehicle maneuver. B & C: Large errors in estimating the intersection location and bounds using geolocation. GPS locations of intersection center sampled over multiple visits at two separate intersections are shown (magenta points). For some instances the locations are estimated outside of the roadway.**

Due to the obvious limitations of primarily using geolocation information for finding and characterizing intersection locations, our solution strategy relies on fusing location information, vehicle speed and the information obtained by processing the scene camera videos. The approximate location of intersections can be interactively selected so that processing can be focused on specific intersection segments. Using computer vision algorithms to process the scene videos, we can obtain a variety of information with relatively high accuracy – such as type of intersection signage (traffic light, or stop sign etc.), traffic density, the maneuver, when the vehicle entered the intersection, and when it exited the intersection. Particularly, precisely determining vehicle's entry point into the intersection is important for the outcomes related to drivers' head scanning behaviors. Our solution framework uses a diverse array of input information to make inferences about driving behaviors at intersections (Figure 3).

The inference framework mentioned in Figure 3 can be considered as a collection of algorithms that take one or multiple inputs and produce/affect a specific outcome or a set of outcomes. The algorithms within this framework could work independently or interdependently. The overall idea is that with the various different kinds of data collected using the in-car recording system, a detailed analysis of the events and corresponding driver behavior should be possible.

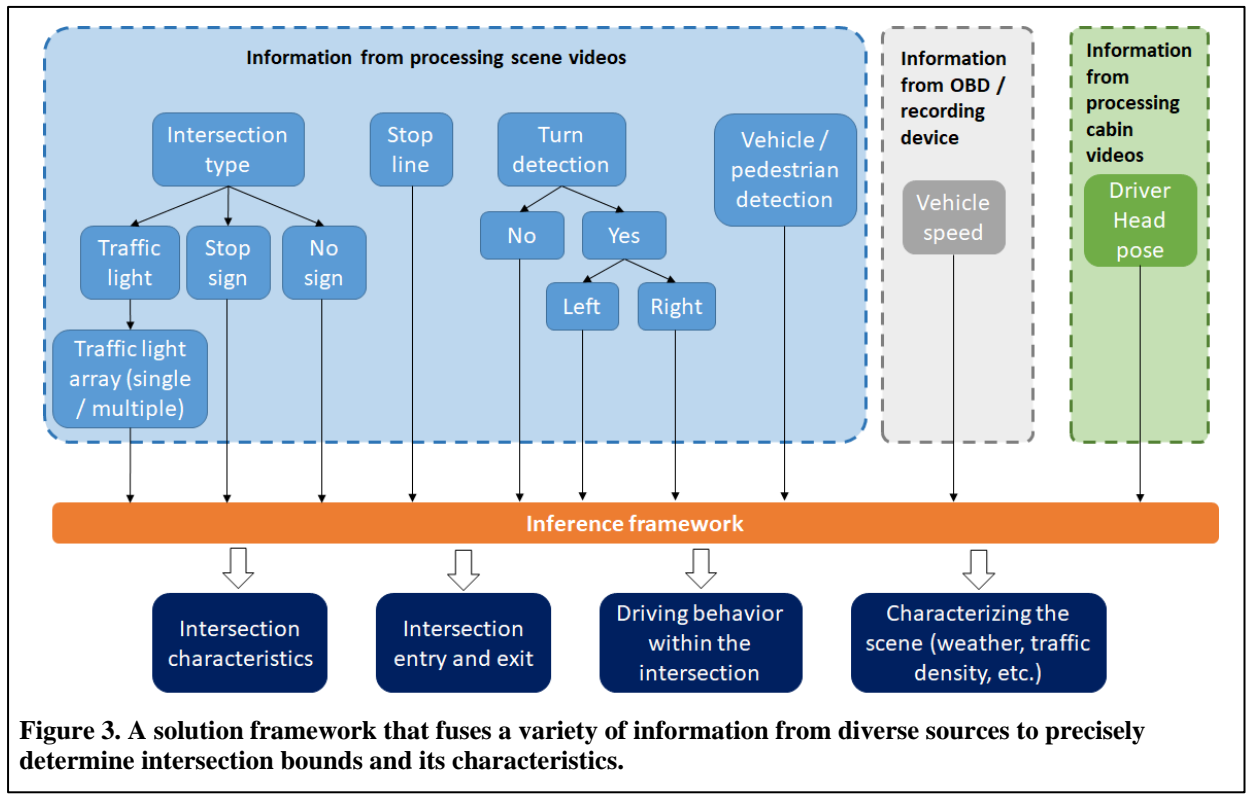

Figure 3. A solution framework that fuses a variety of information from diverse sources to precisely determine intersection bounds and its characteristics.

## 4. Approach

We use a semi-automated pipeline for processing the data collected in naturalistic driving (Figure 4). The processing relies on off-the-shelf as well as custom-developed tools. It is semi-automated because the intersections are selected and marked on the map manually, as it provides a flexible way to focus on drives or locations of specific interest depending on the aims of the investigation. We have designed a custom tool for marking the intersections on the map and clipping the videos associated with the marked intersection locations/instances. Thus, we have a video clip associated with an intersection instance. Further data processing steps are automated. Once the scene and cabin videos are clipped, date-time and speed information embedded in the video frames is extracted. The extracted date-time information (YYYMMDD, HHMMSS) is used for precise synchronization of cabin and scene videos at the frame level (time increments in both the videos are matched frame-by-frame). Cabin video is processed to detect head pose and then head scans are computed from the head position data. The scene video is processed to detect i) scene objects – stop signs, traffic lights, vehicles, and pedestrians, ii) stop line(s), and iii) vehicle self-motion patterns for detection of turns.  Together with the speed information extracted from the video frames (and the distance from the start of the video clip obtained by integrating the speed data), the information from processing the scene videos is input to a rule-based decision making algorithm to determine the vehicle entry and exit points from the intersection, as well as characterizing the intersection scene. Based on the precise intersection bounds, the head scans occurring just before entering the intersection and when within the intersection are extracted. We now describe methodological details of the key data processing steps.

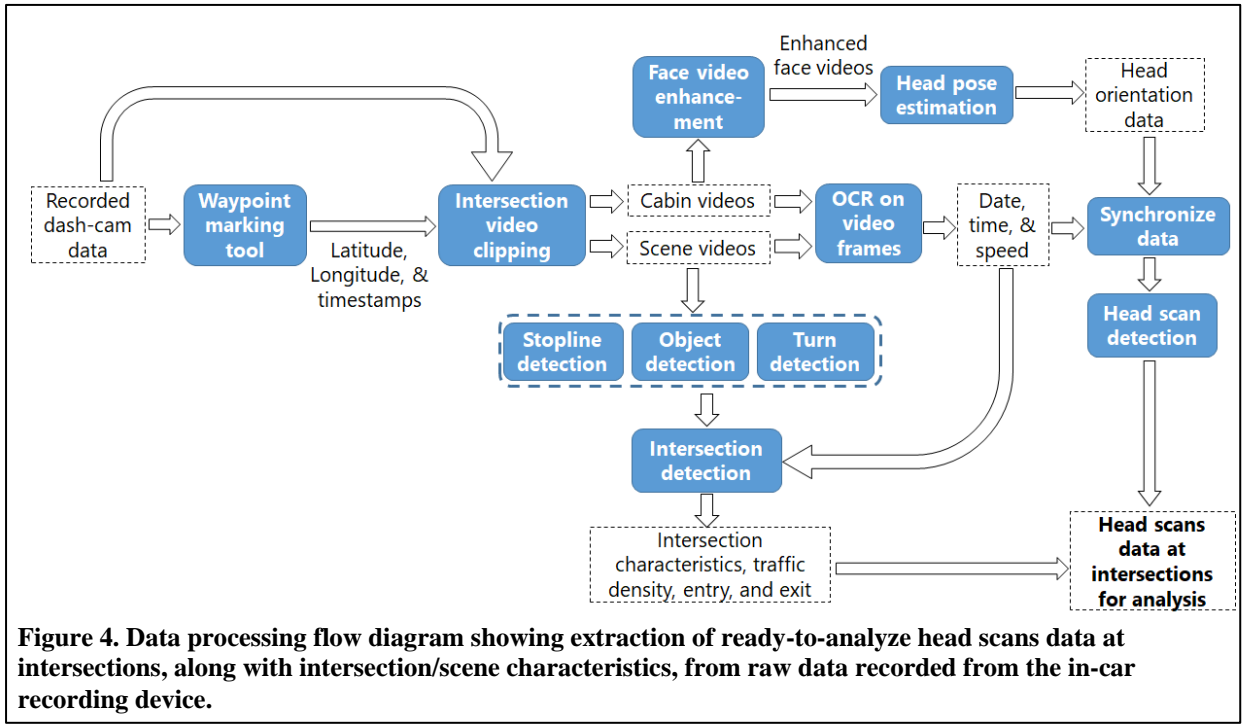

**Figure 4.** Data processing flow diagram showing extraction of ready-to-analyze head scans data at intersections, along with intersection/scene characteristics, from raw data recorded from the in-car recording device.

*4.1 Intersection Selection and Waypoint Marking*
Each trip consists of a set of waypoints where each waypoint represents a timestamp, GPS-latitude, and GPS-longitude, and approximately every two seconds a waypoint is recorded and added to the trip. A custom tool for marking intersections with the help of a human annotator was created. The tool overlaid the waypoints of trips on a map to aid manual marking of the center of each intersection of interest. An early version of the tool also added road signage[8] around intersections, whenever such information was available. However, information related to the signage was often inaccurate which necessitated human oversight and correction, and eventually required automated approaches for detection of intersection signage (described below). The timestamp, latitude, and longitude of each marked intersection were recorded by the tool. For each selected intersection, the waypoint timestamps were used to extract relevant video segments from both cabin and road-facing dashcams. These videos were then synchronized using their timestamps.

*4.2 Stop line Detection*
The stop line is the horizontal line (usually white in color) on the road surface near traffic lights or stop signs that marks the position on the road where the vehicles are expected to stop before proceeding whenever appropriate. Our goal was to determine when the vehicle crossed the stop line, as an important indicator of when the vehicle entered the intersection. For this we first detected all the stop lines present in the given scene video by monitoring image intensity changes within a fixed region of interest (ROI) just in front of the vehicle. Whenever the line is present, there is usually a sharp change in the image intensity. As the vehicle moves, the change in intensity shifts closer to the hood of the vehicle or towards the bottom of the preset ROI. By detecting the sharp intensity change and by monitoring its movement as the vehicle moves, we can determine when the vehicle crossed the stop line.

Real-world situations pose additional challenges. At some intersections, the stop line may not be present, or is faded, broken, washed out due to reflections (in rain), or salted over (in winter). Thus, the contrast of the line may vary based on the road conditions as well as the weather. The stop line could be occluded by the vehicle in front (e.g., if the vehicle with the dashcam was very close to the vehicle ahead). In some cases, there could be zebra stripes near the stop line. In other cases, multiple parallel lines could be present – indicating a pedestrian crossing path without the zebra crossing lines. Also, stop lines are not the only white markings on the road, as white arrows that indicate turning direction for the lanes, and/or text such as "stop" or "turn only" could be present just before the stop line. As one approaches the intersections, presence of these markers could also result in sharp intensity changes in the ROI, and need to be distinguished from the actual stop line.

The ROI for stop line detection, 500 pixels wide and 100 pixels in height, was placed in the scene video frame just above the part of the hood which was visible (Figure 5A). Based on the camera position and the vehicle height, the exact placement of the ROI was adjusted for each vehicle. For each scene video frame, the ROI was cropped, converted to gray scale, and the intensity values were summed column-wise. Since the stop line may not appear exactly horizontal across the cropped ROI, the region was rotated from 4° clockwise to 8° counter-clockwise with a step-size of 1° and the maximum value was selected. If a stop line was present in the ROI, the column-wise sum of the intensity values (normalized) peaks at the row corresponding to the stop line. Peaks were detected using the following criteria: minimum peak height of 0.5, minimum peak prominence (height from the baseline) of 0.17, minimum line width of 2.5 pixels, and maximum line width of 12 pixels. Together these parameters helped filter out spurious detections due to reflections, shadows, or other markers present on the road surface.

Stop lines detected via frame-by-frame processing using the above method were added to a list. Valid stop lines appeared continuously in multiple frames, usually at the top of the ROI and shifting towards the bottom as the vehicle approached the stop line (Figure 5B). This pattern was used to identify real stop lines and filter-out spurious detections. Detections that appeared over at least 5 frames continuously, over a distance of at least 1 meter (distance covered by the vehicle during approach), that first appeared in the top half of the ROI and moved towards the bottom half of the ROI were selected as real stop lines. The final frame in a given stop line sequence was considered as the frame when the vehicle crossed that stop line (with an additional buffer of 1m).

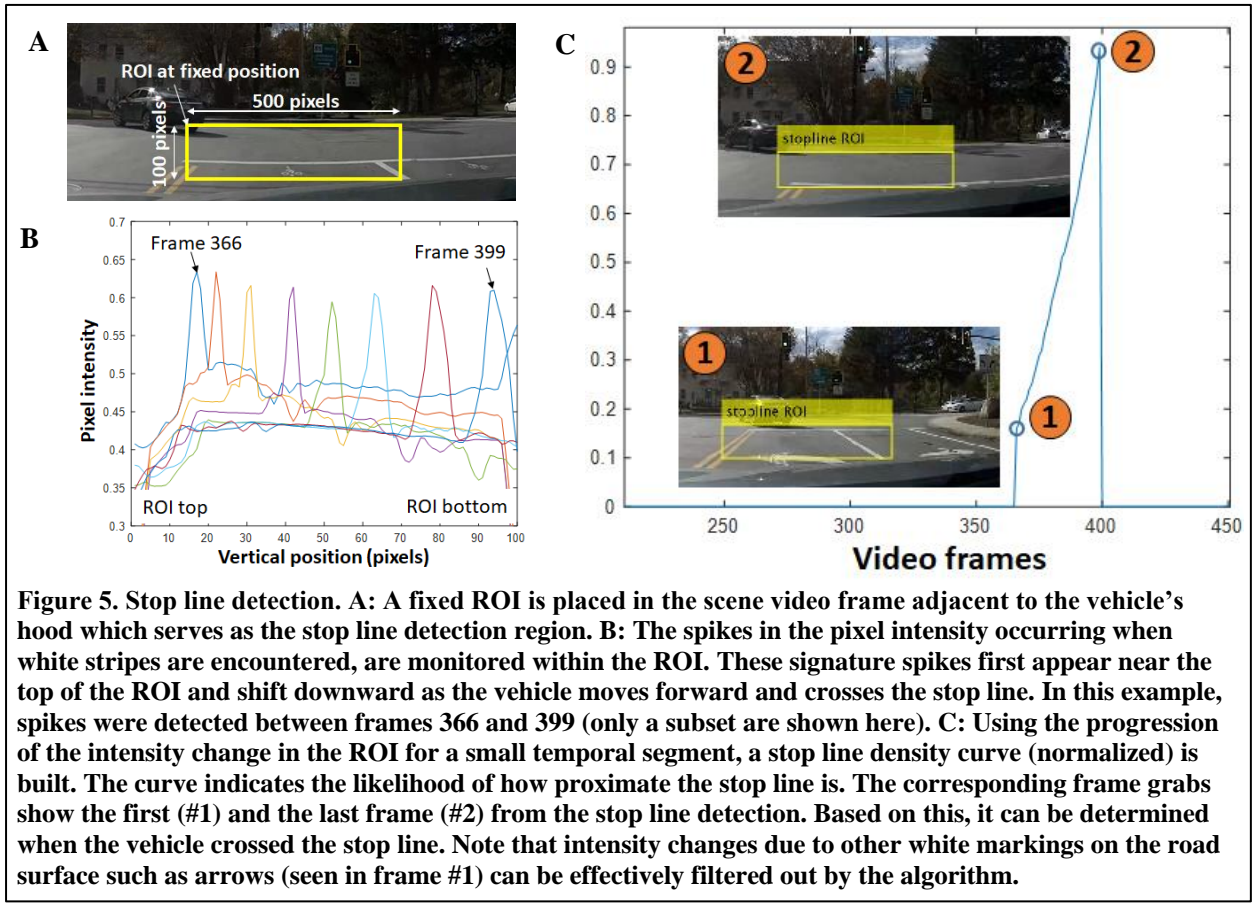

**Figure 5. Stop line detection. A:** A fixed ROI is placed in the scene video frame adjacent to the vehicle's hood which serves as the stop line detection region. **B:** The spikes in the pixel intensity occurring when white stripes are encountered, are monitored within the ROI. These signature spikes first appear near the top of the ROI and shift downward as the vehicle moves forward and crosses the stop line. In this example, spikes were detected between frames 366 and 399 (only a subset are shown here). **C:** Using the progression of the intensity change in the ROI for a small temporal segment, a stop line density curve (normalized) is built. The curve indicates the likelihood of how proximate the stop line is. The corresponding frame grabs show the first (#1) and the last frame (#2) from the stop line detection. Based on this, it can be determined when the vehicle crossed the stop line. Note that intensity changes due to other white markings on the road surface such as arrows (seen in frame #1) can be effectively filtered out by the algorithm.

*4.3 Vehicle Motion/Turn Detection*

Vehicle self-motion was computed based on the image motion between successive scene video frames. Turn detection was based on the idea that self-motion patterns change when the vehicle stops or turns. Particularly, when the vehicle was moving in a straight line (translation along the camera axis without any rotation), the typical radial motion pattern was observed around the focus of expansion (FOE) – with smaller motion magnitudes closer to the FOE that became larger towards the periphery of the image. The FOE was typically close to the center of the image frame. When the vehicle turned, the rotational motion of the vehicle induced a strong image motion component along the horizontal axis including near to the center part of the image frame.

For vehicle motion, the horizontal component of the image motion was more critical than the vertical component (which could theoretically be interesting if the vehicle was passing over uneven terrain, but this was not our case). The bottom part of the 1920×1080 resolution scene video frame was generally occupied by the vehicle's interior and its hood, and this part was excluded from processing the scene videos. The scene image was cropped to 1920×640 pixels and then divided into 12×4 blocks, each of 160×160 pixels. Average motion in the horizontal direction within each block was computed, such that a grid of 12×4 motion vectors were obtained. The motion signal was further summarized in 2 ways: i) median motion of the central 2 columns of the grid, and ii) median motion after summation of the first 5 columns with the mirrored order of the last 5 columns of the grid (i.e., column1+column12, column2+column11, and so on). Both of these values were computed for each scene video frame and then analyzed as a time series. The first value, referred to here as M1, indicates a turn – as it peaks at the middle of the turn when the horizontal motion component is strongest (Figure 6A). The second value, referred to here as M2, also increases during turns, but it is more useful in indicating when the vehicle is stopped (absolute minimum).

Both M1 and M2 signals were smoothed with a moving median window of 15 frames (30 frames if the signal was excessively noisy, for example when the windshield wipers were on during rain). For turn detection, peaks of the normalized M1 signal were determined with the following parameters: minimum peak height of 0.25, minimum peak prominence of 0.175, and with a spacing > 3seconds between successive peaks. The detected turn candidates were further pruned based on the minimum distance spanned during the turn (4.5 meters) and minimum speed during the peak turn (5 miles per hour). These criteria helped in separating actual turns at intersections from road curves or other driving maneuvers. For the remaining peaks in the M1 signal that were likely to be turns at intersections, the start and end of each turn were determined from the extent of the peak signal from the baseline, which was based on the location and prominence of the detected peaks (Figure 6B). The sign of the detected peaks indicated whether it was a left (+ve) or right (-ve) turn. When no peaks were detected, it was assumed that the vehicle went straight through the intersection. Vehicle speed and the M2 motion signal were used to determine whether the vehicle stopped anytime during the video.

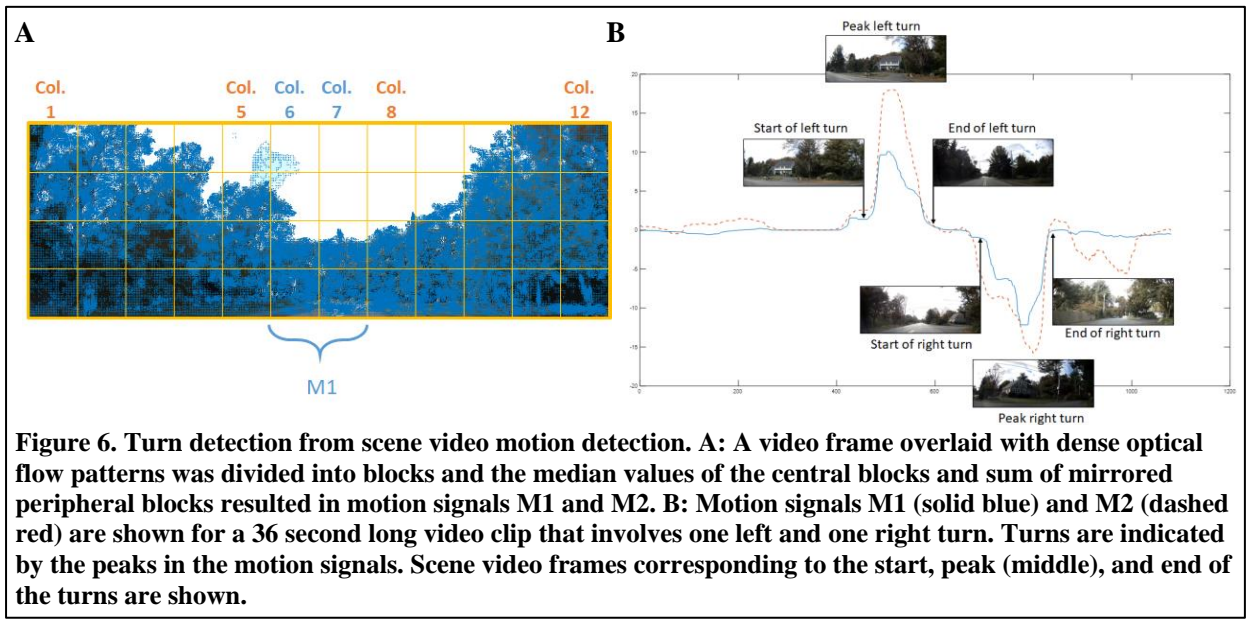

**Figure 6. Turn detection from scene video motion detection. A: A video frame overlaid with dense optical flow patterns was divided into blocks and the median values of the central blocks and sum of mirrored peripheral blocks resulted in motion signals M1 and M2. B: Motion signals M1 (solid blue) and M2 (dashed red) are shown for a 36 second long video clip that involves one left and one right turn. Turns are indicated by the peaks in the motion signals. Scene video frames corresponding to the start, peak (middle), and end of the turns are shown.**

*4.4 Object Detection & Scene Characterization*
The YOLO object detection model[9] trained on the COCO dataset[10] was used for detecting scene objects. Each scene frame was input to the object detection model and the output was a list of objects (bounding box, class label, confidence) per frame. The COCO dataset was trained on more than 90 common object classes, but only a few are of relevance in a traffic scenario: vehicles (car, truck, bus, bicycle, and motorcycle), stop sign, traffic light, and person (pedestrian). Any detections corresponding to the rest of the object classes were ignored. For any given scene video, a list of detected objects was obtained. There could be multiple objects within a single video frame, while some frames may have no detections. Reliability of object detection was determined based on the temporal trends of object appearance.

Based on the list of detected scene objects, we further characterized the intersection scenario. First, we determined the kind of signage at the intersection. In our case, there were 3 types of intersection signage –stop sign, traffic light, or neither of these (none). We checked for the presence of "stop sign" or "traffic light" object class label within the object detection list (should be present for more than 20 frames). If neither of the two was present, then the intersection type was none.

For stop signs, we restricted to detections only in the right half of the frame that corresponded to the right side of the vehicle. Also, a maximum height of 150 pixels was enforced on the detected bounding box to remove false positives. Depending on the roadway geometry and external conditions, the object detection model could start detecting stop signs when the vehicle was many meters away. The bounding box size increased with the vehicle's approach until the sign moved out of the field of view of the scene camera. This fact was used to cluster frames during which stop sign appeared in the video and a density function was created based on the area of the stop sign bounding box in the image (Figure 7A).

For traffic lights, detection was restricted to the central 70% of the frame, with a maximum bounding box height of 85 pixels. Traffic lights need to be processed somewhat differently than stop signs, because unlike a stop sign, there could be single (Figure 7B) or multiple arrays of traffic lights visible from the scene camera for the given intersection (Figure 7C). The YOLO object detection model detects the traffic lights regardless of their orientation, so there could be multiple traffic light bounding boxes simultaneously present that correspond to the same intersection. For example, when approaching a 4-way intersection, there could be one traffic light array facing the vehicle, one with its back to the vehicle, and 2 appearing sideways. Depending on the maneuver, the bounding box areas of one array could increase or decrease as the vehicle passes through the intersection. Therefore, the largest bounding box per frame was selected to construct the traffic light density function, which could show multiple peaks, depending on the number of traffic light arrays present.

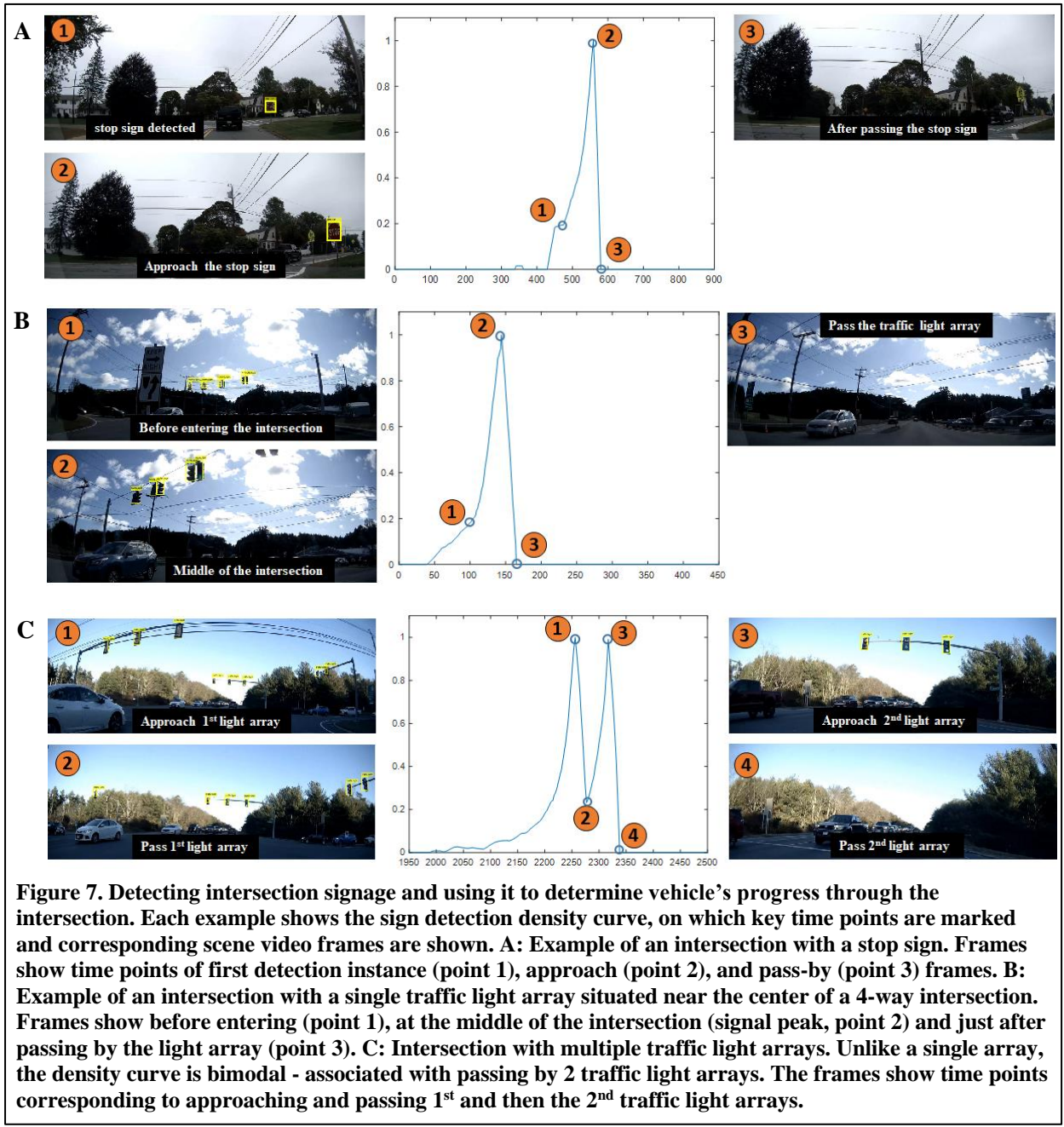

**Figure 7. Detecting intersection signage and using it to determine vehicle's progress through the intersection. Each example shows the sign detection density curve, on which key time points are marked and corresponding scene video frames are shown. A: Example of an intersection with a stop sign. Frames show time points of first detection instance (point 1), approach (point 2), and pass-by (point 3) frames. B: Example of an intersection with a single traffic light array situated near the center of a 4-way intersection. Frames show before entering (point 1), at the middle of the intersection (signal peak, point 2) and just after passing by the light array (point 3). C: Intersection with multiple traffic light arrays. Unlike a single array, the density curve is bimodal - associated with passing by 2 traffic light arrays. The frames show time points corresponding to approaching and passing 1st and then the 2nd traffic light arrays.**

Another intersection scene characteristic was traffic density, which was defined as the average number of vehicle bounding boxes present per second (moving average of the vehicle bounding box count over 1 second duration). This number included all detected vehicles, including those not on the roadway. Also, this was not an indicator of how many unique vehicles were seen in the video. Although this could be computed by tracking the vehicles over time as they enter and exit the scene, it was beyond the scope of the current work. The vehicle bounding boxes were further classified based on their aspect ratio into those that were part of cross-traffic (their bounding boxes tended to be rectangular vs. those in the same direction of travel typically were more square shaped).

*4.5 Intersection Bounds Detection*

The algorithm for intersection bounds detection works by combining information obtained from processing the scene videos. We assumed that only one intersection was present in the given scene video (so the case of multiple intersections close together is beyond the scope of this work). For each video clip of an intersection, the first step was to determine the type of intersection signage and the maneuver(s) made. Since each scene video element was detected or processed independently initially, the algorithm then associated, based on proximity, the detected stop line(s) and the turn maneuvers with the intersection. This was important when there were multiple turns or multiple stop lines detected in the video, but only one of them was truly associated with the intersection. After that, the length of halt (if any) near the intersection was determined. Using these various pieces of information, we determined the entry and exit points based on the intersection entry and exit criteria defined below.

Precise entry and exit from the intersection were determined by combining, in a hierarchical rule-based procedure, the information obtained from processing the scene videos: vehicle speed (and distance derived from the speed), stop line crossing, type of intersection signage (stop sign, traffic light, or none), and maneuver (right turn, left turn, or straight). Intersection entry was defined using objective criteria. For intersections with stop signs, the frame when the stop sign was first out of the camera field of view was considered as the point of entry into the intersection. Because not all stop sign intersections have a stop line (especially in Massachusetts), crossing the stop line at stop signs was not considered in the current version of our algorithm.

For traffic light intersections, the situation was complicated by intersection geometry, the status of traffic lights (green or red) and the traffic in front of the vehicle. It was possible that the vehicle just crossed the stop line and came to a halt at a light, or it was possible that they waited behind the stop line. Therefore, entry into the intersection was determined as the point of crossing the stop line or when the vehicle first started moving after a halt close to the stop line, whichever was later (Figure 8). Sometimes the stop line could not be detected reliably because of the various factors noted previously. In such situations, the criteria were modified to consider when the vehicle passed by the traffic lights (when the traffic lights go out of the field of view of the camera). In intersections with a single traffic light array, the entry point was fixed as the point when the vehicle was 15 meters from where it passed the traffic lights array (because the array was likely in the middle of the intersection). In the case of multiple arrays of traffic lights, the entry point was when the vehicle passed the array closest to the vehicle approach (seen as the local minima between the two peaks of the density function).

Compared to intersection entry, determining the exit from an intersection was somewhat less objective because there was a lack of consistently clear landmarks to use as reference points. Whenever turning maneuvers were involved, we chose completion of the turn as the point of exit from the intersection. When no turns were involved, then the exit point was chosen as a point when the vehicle traveled 2.5 m after passing the last traffic light array when multiple light arrays were involved, or 15 m for a single light array (because it was likely that the single array was in the middle of the road and the vehicle had to travel a longer distance to completely exit the intersection). In the case of stop signs without turns the exit was estimated as the point when the vehicle had traveled 30 m from the entry point as there were no other indicators to mark the exit. The pre-set distances used in our approach were based on observations (as there is no single nationwide standard for intersection design). In actual analysis, these could be changed based on the general driving location (for example, Northeast US vs. the Midwest).

If the intersection did not have a stop sign or a traffic light (i.e., it was classified as "none"), then the only objective way to mark the entry and exit from the intersection was based on the start and end of the turn maneuver. If there was no turn, then it was not possible to identify the entrance and exit points. In this paper, we excluded such situations (no sign and going straight).

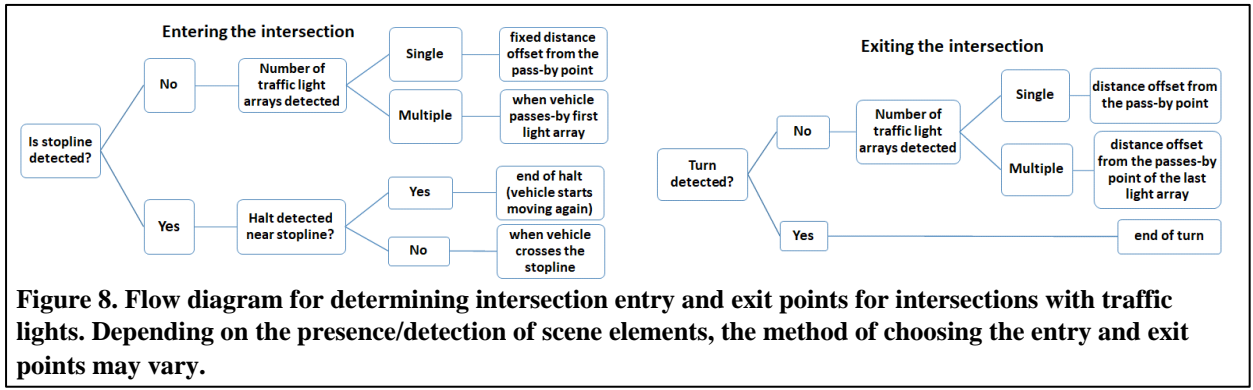

**Figure 8.** Flow diagram for determining intersection entry and exit points for intersections with traffic lights. Depending on the presence/detection of scene elements, the method of choosing the entry and exit points may vary.

*4.6 Cabin Video Processing*

The cabin video was processed to extract head pose. A key aspect was that the position of the cabin camera was fixed and did not change with respect to the driver. We first cropped the cabin video around the driver's head (400×300 pixels) and then performed contrast-limited adaptive histogram equalization to enhance the face image of the driver.[11] Contrast enhancement was necessary because illumination can change drastically when driving (sunlight, glare or shadow, low light conditions etc.). First, the cabin video was cropped and converted to LAB color space.

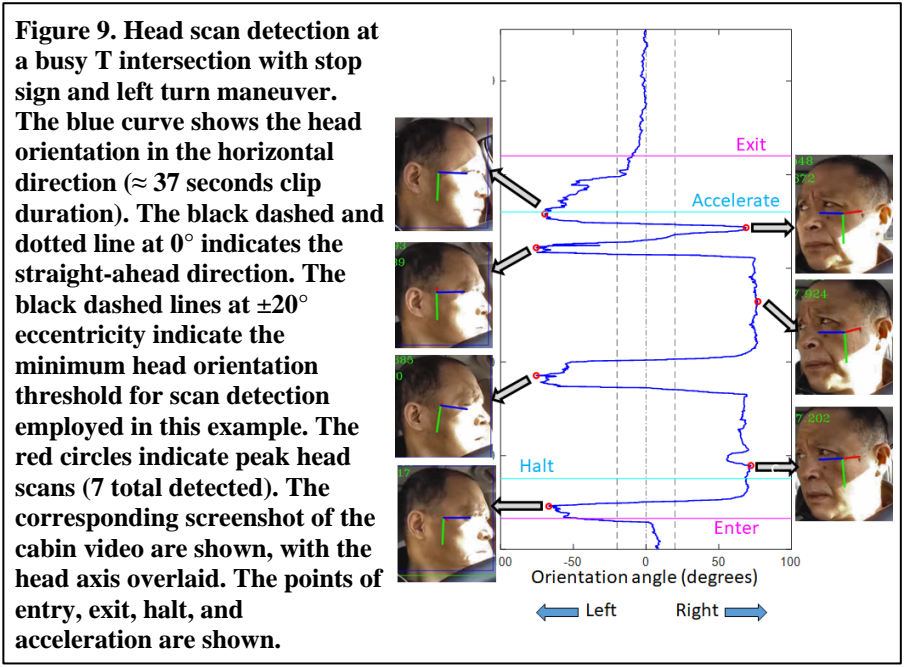

**Figure 9.** Head scan detection at a busy T intersection with stop sign and left turn maneuver. The blue curve shows the head orientation in the horizontal direction (≈ 37 seconds clip duration). The black dashed and dotted line at 0° indicates the straight-ahead direction. The black dashed lines at ±20° eccentricity indicate the minimum head orientation threshold for scan detection employed in this example. The red circles indicate peak head scans (7 total detected). The corresponding screenshot of the cabin video are shown, with the head axis overlaid. The points of entry, exit, halt, and acceleration are shown.

Enhancement was performed on the L channel by dividing it into an 8×8 grid of tiles, computing histograms (256) within each tile, and equalizing them over the full grayscale range (0-255) assuming Rayleigh distribution ($\alpha$=0.4), with a contrast enhancement limit of 0.005. The enhanced image was converted back to RGB space.

The head pose detection algorithm was then applied to the enhanced image. There are open source deep learning models available for head pose detection. However, we developed our own method because the head orientation at intersections can be large (> 90° yaw rotation relative to the camera) and the current head pose detection models are not able to reliably handle situations with such large head turns. Our deep learning algorithm involved 2 components: precise head detection and wide-angle head-pose estimation. The head was localized by using a custom trained YoloV6 model that can reliably detect human heads wearing hats and/or glasses across a wide range of head poses. Once localized, the head pose was estimated by using a custom fine-tuned wide range head pose estimation model, 6DRepNet360.[12] The fine tuning was conducted based on a combination of the CMU Panoptic dataset[13], 300W-LP[14], and AFWL2000.[15]

Head pose for each video frame was available as the yaw, pitch, and roll angles of the head with respect to the straight-ahead direction of the driver. Head scans were detected using an algorithm that was previously evaluated in the context of walking mobility.[16] A head scan was defined as the movement of the head away from the straight-ahead direction, either towards left or right side, before returning back towards the straight-ahead direction (Figure 9). The head scan magnitude or peak head position was the farthest the head moved away from the straight-ahead direction for a given scan. Because small head movements do not play an important role in scanning a large field of view (e.g., 180°) at intersections,[17] head scans were defined as those where the peak head position exceeded a minimum threshold magnitude. In our implementation, we set the threshold at 20° from the straight-ahead direction.[18] Thus, a head scan was detected when the head yaw position was > 20° for at least 5 frames before returning below 20°. In the case of compound scans, where the head moved away before fully returning back within 20° of the straight-ahead direction, the peak position was retained as the scan magnitude. Since head scans are computed for the entire video clip, relevant scans can be selected as per any preset criteria as needed for the analysis.

## 5. Evaluation Results

Our methods described above were evaluated in 3 experiments that measured the accuracy of intersection characterization, intersection bounds detection, and head pose estimation.

### *5.1 Intersection Detection*

Algorithmically estimated intersection characteristics and bounds (entry and exit points) were compared with the ground truth obtained via manual annotation of the scene videos.

5.1.1. Characteristics of the naturalistic driving data set used
A total of 190 video clips of intersections from 3 vehicles (2 from MA, 1 from CA) were annotated. Two of the drivers had homonymous visual field loss and one had no visual field loss. The driving locations were a mix of urban and suburban settings and were mostly during the daytime (some clips at dawn or dusk). Intersection video clips were extracted to cover close to 200 meters distance (±100 meters from the middle of the intersection). The median duration of each intersection video clip was 23[18-37] seconds, and the median distance from the start of the clip to intersection entry was 88[77-101] meters. The median intersection span was 36[30-42] meters with the drivers spending 5[3.2-6.6] seconds within an intersection (from entry to exit).

5.1.2 Video Annotation and Outcome Measures
Intersection entry and exit were manually marked by three researchers annotating the respective frame numbers in the scene videos as ground truth. A custom visualization and annotation software was developed for playing the scene videos frame-by-frame. The criteria described in section 4.5 above were used for manual marking of the intersection entry points depending on the type of intersection. Marking exit frames was based on the best judgment of the reviewer (because, in contrast to the entry point, there were fewer consistently available landmarks to specify criteria to determine a vehicle's exit from the intersection). Since the speed of the vehicle was known, the ground truth distance of the vehicle entry and exit from the intersection from the beginning of the clip could be computed. For each test intersection clip, our algorithm predicted the entry and exit frames along with the corresponding distances. We used the following outcome measures: i) comparing estimated entry time (from the start of the clip) with ground truth, ii) comparing estimated entry distance (from the start of the clip) with ground truth, iii) comparing estimated overlap (exit – entry) of intersection bounds with the ground truth bounds (using dice coefficient), and iv) comparing the number of head scans detected within actual and estimated intersection bounds. Given the intersection exit was somewhat arbitrarily marked and also because the vehicle may be stopped for a long time at the light before entering the intersection, we also computed the difference in the number of head scans within a window of ±5 seconds around the estimated and ground truth intersection entry point. This severed as an alternative way to determine if the head scan count just prior to and just after entering the intersection was comparable between automated and manually annotated intersection entry point. When reviewing the clip, type of intersection signage (stop sign, traffic light, or none) and the maneuver (left, right, or straight) were annotated and were compared with the estimated values. The intersection bounds accuracy measures were computed across all intersections and for each intersection type and maneuver performed. Analysis of head scanning behaviors (scan frequency, magnitude, etc.) is out of the scope of this report and will be presented in a separate manuscript.

5.1.3 Accuracy of Intersection Characterization

The data set included 135(71%) 4-way, 41(22%) T, and 14(7%) Y intersections. There were 30(16%) stop signs, 118(62%) traffic lights, and 42(22%) intersections without any signs. The drivers made 46(24%) left turns and 54(28%) right turns, while maneuvering straight through the intersection in 90(48%) cases. The processing algorithm was able to correctly detect the type of intersection and type of maneuver in 100% and 94% of instances, respectively. The overall median absolute difference [IQR] between the estimated and ground truth entry time was 0.2[0.1-0.54] seconds, with an RMSE of 1.7 seconds. The entry time difference was within a 1 second margin for 91% of the instances (Figure 10A). The entry time was off by more than 5 seconds in only 2 instances. The median absolute distance between the estimated and actual entry point was 1.1[0.4-4.9] meters, with an RMSE of 6.0 meters. The estimated entry point was within 10m of the ground truth in 93% of the instances (Figure 10B). The median overlap between ground truth and estimated intersection bounds was 0.88[0.82-0.93], with 86% of intersection bounds recording an overlap > 0.75 and only 3 cases with overlap<0.5 (Figure 10C). The number of driver head scans counted within the ground truth and estimated intersection bounds were different for 16 (8.4%) intersections, with a single scan difference in 15 cases and 2 scan difference in 1 case. The median difference in the entry time, entry distance, and overlap for the 16 instances of differing head scans were 0.63[0.22-1.56] seconds, 0.64[0.2-6.9] meters, and 0.82[0.71-0.9], respectively. Head scans were different for 15 (8%) intersections when comparing the ±5 seconds window (ground truth vs. estimated), with 13 intersections recording a difference of a single head scan.

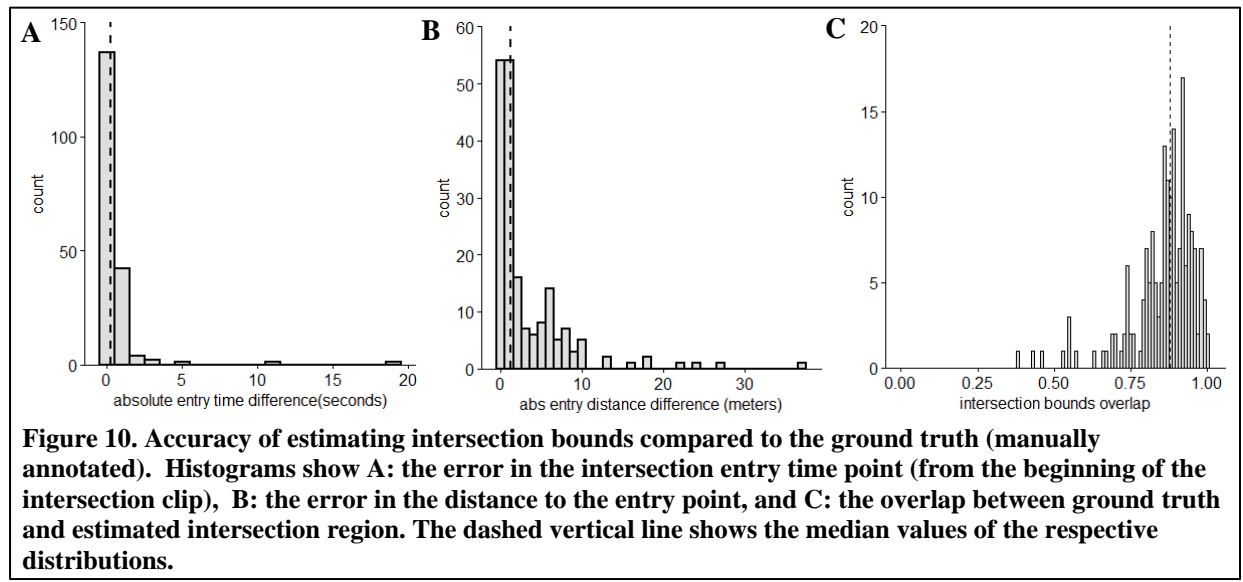

Figure 10. Accuracy of estimating intersection bounds compared to the ground truth (manually annotated). Histograms show A: the error in the intersection entry time point (from the beginning of the intersection clip), B: the error in the distance to the entry point, and C: the overlap between ground truth and estimated intersection region. The dashed vertical line shows the median values of the respective distributions.

The intersection bounds detection accuracy outcomes varied by the type of intersection and the maneuver performed (Table 1). The entry point time error was significantly different among the 3 signage types ($\chi^2$= 34.2, df=2, p<0.001), with a significantly larger difference between "None" compared to "Stop sign" (p<0.001) and "Traffic light" (p<0.001). Similarly, the entry point distance errors were significantly different among the signage types ($\chi^2$= 52.6, df=2, p<0.001), with a significant pairwise difference between "None" and both "Stop sign" (p<0.001) and "Traffic light" (p<0.001) types. The intersection segment overlap was significantly higher for "Stop sign" compared to the other 2 categories, and the overlap was significantly lower in "None" compared to "Traffic light". There was a significant effect of maneuvers performed on the bounds detection, with a right turn maneuver being the worst (p<0.001) among the 3 maneuver categories for entry point time and distance errors. Overlap was significantly higher for intersections with left turns (p<0.001) compared to straight or right turns.

Table 1: Intersection bounds detection results for different intersection types and maneuvers.

| Variable | Categories | Count | Entry point time error (seconds) | Entry point distance error (meters) | Intersection segment overlap |
|---|---|---|---|---|---|
| Signage type | None | 42 | 0.70 [0.43 – 0.92] | 6.8 [4.9 – 8.5] | 0.83 [0.74 – 0.87] |
| | Stop sign | 30 | 0.22 [0.11 – 0.39] | 0.9 [0.4 – 2.0] | 0.95 [0.90 – 0.97] |
| | Traffic light | 118 | 0.13 [0.07 – 0.33] | 0.8 [0.2 – 1.6] | 0.88 [0.82 – 0.92] |
| Maneuver performed | Left | 46 | 0.23 [0.10 – 0.48] | 0.5 [0.1 – 1.3] | 0.94 [0.89 – 0.97] |
| | Right | 54 | 0.55 [0.21 – 0.80] | 4.9 [1.0 – 7.5] | 0.86 [0.81 – 0.92] |
| | Straight | 90 | 0.13 [0.07 – 0.30] | 1.1 [0.5 – 2.5] | 0.87 [0.81 – 0.90] |

Values in brackets are $[25^{th} - 75^{th}$ percentile].

### 5.1.4 Error Cases

While there were no misidentifications of intersection signage, there were 11 instances (6%) of misidentification of maneuver. These were mostly because of curving roads near the intersection that were identified as a turn, but were labeled as a straight maneuver in manual reviewing. There were 2 cases where the algorithm failed to detect intersection bounds completely. They were intersections without signage ("None') where the drivers made a left turn. The reason for the detection failure was turn detection failure – the turn span did not meet the criteria in one instance whereas in the other instance the motion signal was not strong enough to be distinguished from a curve in the road. There were also some instances where large errors in intersection entry point estimation were seen. We defined large errors as intersection entry time > 1 second or intersection entry distance > 10 meters. There were 18 such cases (9%). In some of these large error instances, there was a delay of the order of many seconds in the odometer reading to be updated in the scene frame, which affected the estimation of vehicle entry point (as this happened at traffic lights). In other instances, large turns were negotiated in a piecemeal manner. Thus, the motion patterns did not provide a sufficiently strong signal to classify it as a turn. Failure to detect a stop line was one of the main sources of error. This happened because in some intersections, the stop line was either faded, or broken, or there was glare, or rain distorted visibility. Some errors were due to unusual geometry of the intersection, with uncommon placement of the signs or traffic lights.

### *5.2 Head Pose Accuracy Measurement*

For the future implementation of our methods to evaluate head scanning at intersections, we were mostly interested in lateral head rotations (yaw movements). Even though the head pose algorithm outputs the yaw, pitch, and roll angles of the head orientation, our experimental evaluation below only focused on the yaw angle measurement of the head pose.

### 5.2.1. Experiment Methods

The experiment for head pose accuracy measurement was performed in a 20×20 feet (6×6 meters) room. Three subjects with normal vision turned their heads over a wide angular range of ≈ ±135°. The subjects were seated approximately in the middle of the room and targets were attached to the walls around the seating location. One of the targets was close to the straight-ahead direction (or the 0° reference) and 7 targets each were placed on the left and right with increasing eccentricity from the center target. The subjects wore a head mounted laser pointer strapped to the center of their forehead (middle of the two eyes) (Figure 11A) to help align their head with the targets located on the walls around the room. The subjects first aligned their head with the center target (straight ahead direction or 0°), and then successively turned and aligned with targets on the right side till they reached the most extreme target. Then they reversed the order and revisited each target on the right, passing the center, and continued aligning with the targets on the left side all the way through, before returning to the center target. Subjects raised their hand to indicate that they had aligned with the target. Each alignment event lasted about 1 to 3 seconds. A video camera (iPhone) set up on a tripod was placed around 28 inches (71 cm) away from the subject's face that recorded the full sequence of the head turns. The video was processed by the head pose detection algorithm. Video frames associated with raised hand gestures at each target alignment were extracted and the corresponding yaw angles were obtained. Ground truth was computed based on the known positions of the targets with respect to the subject's location. Since the target alignment was known beforehand, the estimated yaw angles could be compared with the ground truth.

5.2.2 Head Pose Estimation Results

The head pose algorithm was able to track head successfully for angles > 90° with respect to the straight ahead viewing direction (Figure 11A). Compared to the ground truth, the estimated yaw pose was highly consistent with the ground truth for all 3 subjects, as indicated by the slope and $R^2$ of the trend line being very close to 1 (Figure 11B). The mean absolute error (±standard deviation) across all 3 subjects and across all target locations was 6.75±5.0°. The root mean squared error (RMSE) was 8.4°. Given that our primary goal is to detect large angle head rotations at intersections, the magnitude of the error is acceptable. For context, the error is roughly equivalent to the half angle spanned by a typical road lane 10 meters (33feet) away from the driver's position. There was no significant difference in the errors between the 3 subjects. The error did not increase significantly with the increasing target eccentricity (no significant correlation).

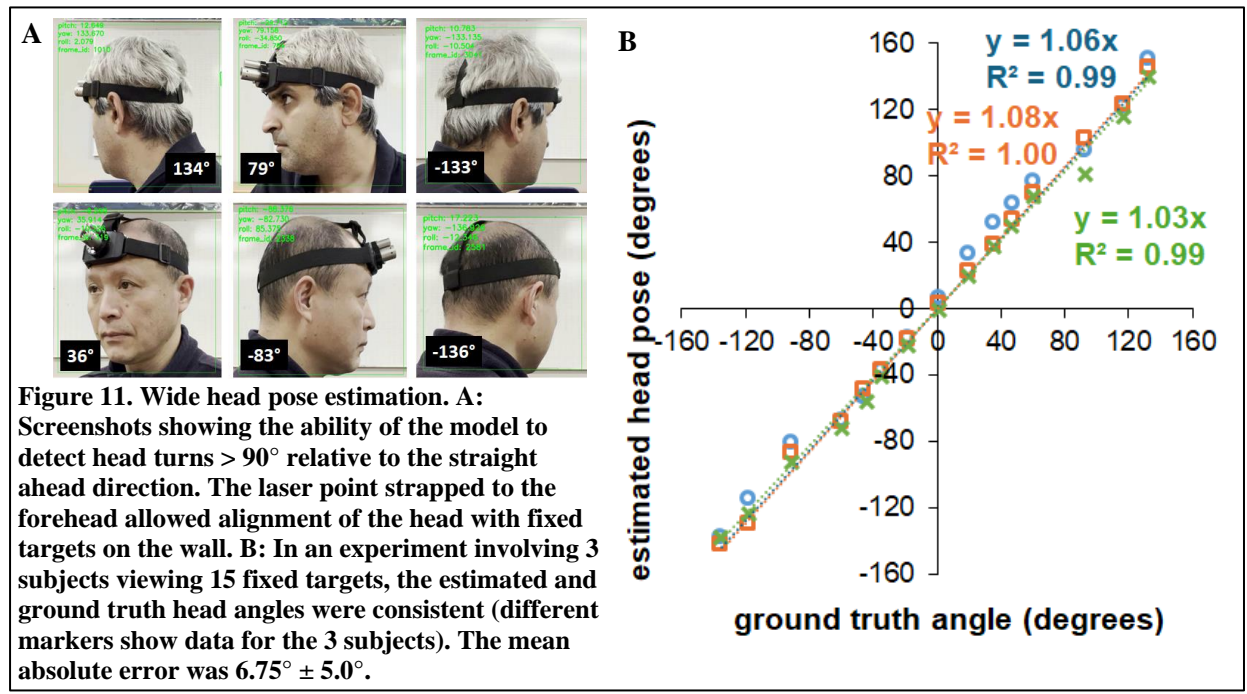

Figure 11. Wide head pose estimation. A: Screenshots showing the ability of the model to detect head turns > 90° relative to the straight ahead direction. The laser point strapped to the forehead allowed alignment of the head with fixed targets on the wall. B: In an experiment involving 3 subjects viewing 15 fixed targets, the estimated and ground truth head angles were consistent (different markers show data for the 3 subjects). The mean absolute error was 6.75° ± 5.0°.

## 6. Discussion

We describe methodological details of collecting and processing of naturalistic driving data with the goal of studying head scanning behaviors of drivers with and without visual field loss at intersections. Because vehicle speed and location information (GPS data) were not sufficient by themselves to accurately determine the vehicle's progress through an intersection, we developed custom methods to characterize the intersection scenario in detail by processing the scene videos. By fusing vehicle speed, location, and scene video processing data, the type of intersection, maneuver performed, as well as precise entry and exit points of the vehicle from the intersection could be determined. Our evaluation found that the algorithm was able to determine the intersection bounds with high accuracy. With the help of our custom developed wide head pose detection algorithm, we were able to extract head scans at intersections in naturalistic driving data.

The automated detection and characterization of intersections makes it feasible to efficiently process large amounts of data without the need for time-consuming manual review (which would be impossible for thousands of intersections), opening up the possibility for detailed analysis of scanning behaviors and factors affecting scanning across several months of recorded driving data.  The precision of the automated detection of the intersection entry point is pivotal to enabling analysis of scanning behaviors across many diverse intersections (i.e. it provides a common reference point across intersection types and maneuvers).

In our evaluation of intersection detection, intersections without signage had the highest errors in terms of entry point and distance, as well as the lowest overlap between estimated bounds and ground truth. Since signage detection is a robust way to determine the intersection entry, absence of signage means we only rely on turn detection. Turn initiation and completion are somewhat more subjective, therefore there may be more arbitrariness to the ground truth annotation, which may lead to seemingly larger errors. When analyzing the errors based on the maneuver performed, right turns seemed to have larger errors because in our sample right turn maneuvers coincided with 81% of intersections without signage. Annotating the intersection exit point is always arbitrary regardless of the type of intersection; therefore, we see relatively smaller overlap of intersection bounds even when the entry point is estimated with a reasonable accuracy (for example, ≈40% of intersections with entry error smaller than the median value had less than the median overlap – this indicates that the exit annotation could have contributed to the overlap error).

There are some limitations of the studies presented in this report. First, we did not evaluate some of the components of intersection scene understanding such as traffic density detection, pedestrian detection, and driving condition determination (adverse weather etc.). This will be future work. Second, our evaluation dataset was limited to video clips with only one intersection each. In actual driving, back-to-back intersections may be present. This is not an issue with the algorithm which can detect multiple intersections, but because only one intersection was annotated per clip, ground truth matching was a problem.  Third, we did not include intersections without signage and without a turn maneuver (i.e., no-sign, no-turn cases). Finally, our processing is currently restricted to daytime videos (including those captured during twilight). Processing nighttime videos would require some additional steps to handle glare and low light regions of the scene video.  Such challenging cases could be handled with more advanced scene understanding functionalities (such as identifying different traffic signs or using GPS information better), which is future work. Finally, we did not include any analysis of head scan behavior, which will be addressed in a separate study.

**Acknowledgements:**
Jina Yi for help with manual marking of intersections and annotation of video clips.

**Funding:**
NIH grant R01-EY025677 (ARB)